\documentclass[lettersize,journal]{IEEEtran}
\usepackage{amsmath,amsfonts}
\usepackage{algorithmic}
\usepackage{algorithm}
\usepackage{array}
\usepackage[caption=false,font=normalsize,labelfont=sf,textfont=sf]{subfig}
\usepackage{textcomp}
\usepackage{stfloats}
\usepackage{url}
\usepackage{verbatim}
\usepackage{graphicx}
\usepackage{cite}
 \newcommand{\textapprox}{\raisebox{0.5ex}{\texttildelow}}
 
 \usepackage{multirow}
 \usepackage{threeparttable}

\usepackage{pifont}
\newcommand{\cmark}{\ding{51}}%
\newcommand{\xmark}{\ding{55}}%

\usepackage{amsmath}
\usepackage{multirow}
\usepackage{makecell}
\usepackage{array}

\usepackage{xcolor}

\newcommand{\toprule}{\hline}
\newcommand{\midrule}{\hline}
\newcommand{\bottomrule}{\hline}

\makeatletter
 \let\old@ps@headings\ps@headings
 \let\old@ps@IEEEtitlepagestyle\ps@IEEEtitlepagestyle
 \def\confheader#1{%
 \def\ps@IEEEtitlepagestyle{%
 \old@ps@IEEEtitlepagestyle%
 \def\@oddhead{\strut\hfill#1\hfill\strut}%
 \def\@evenhead{\strut\hfill#1\hfill\strut}%
 }%
 \ps@headings%
 }
 \makeatother

\confheader{%
 \small{Accepted for publication in IEEE Transactions on Computer-Aided Design of Integrated Circuits and Systems. DOI: 10.1109/TCAD.2022.3212645}
 }

\begin{document}

\bstctlcite{IEEEexample:BSTcontrol}

\title{AdaPT: Fast Emulation of Approximate DNN Accelerators in PyTorch}

\author{Dimitrios Danopoulos,
Georgios Zervakis,
Kostas~Siozios, ~\IEEEmembership{Senior Member,~IEEE,}, Dimitrios~Soudris,~\IEEEmembership{Member,~IEEE,}
and J{\"o}rg~Henkel,~\IEEEmembership{Fellow,~IEEE}%
\thanks{Manuscript received April 9, 2022, revised July 10, 2022, accepted 21 September, 2022. (\textit{Corresponding author: Dimitrios Danopoulos, e-mail:dimdano@microlab.ntua.gr}).}%
\thanks{This work has been supported in parts by the E.C. funded program SERRANO under H2020 Grant Agreement No: 101017168 and in parts by the German Research Foundation (DFG) project ``ACCROSS'' HE 2343/16-1.}
\thanks{D.~Danopoulos and D.~Soudris are with the School of Electrical and Computer Engineering, National Technical University of Athens, Athens 15780, Greece.}%
\thanks{K.~Siozios is with the Department of Physics, Aristotle University of Thessaloniki, Thessaloniki 54124, Greece.}%
\thanks{G. Zervakis and J. Henkel are with the Chair for Embedded Systems at Karlsruhe Institute of Technology, Karlsruhe 76131, Germany.}%
}



\maketitle

\begin{abstract}
Current state-of-the-art employs approximate multipliers to address the highly increased power demands of DNN accelerators. However, evaluating the accuracy of approximate DNNs is cumbersome due to the lack of adequate support for approximate arithmetic in DNN frameworks. We address this inefficiency by presenting AdaPT, a fast emulation framework that extends PyTorch to support approximate inference as well as approximation-aware retraining. AdaPT can be seamlessly deployed and is compatible with the most DNNs. We evaluate the framework on several DNN models and application fields including CNNs, LSTMs, and GANs for a number of approximate multipliers with distinct bitwidth values. The results show substantial error recovery from approximate retraining and reduced inference time up to $53.9 \times $ with respect to the baseline approximate implementation. 
\end{abstract}


\section{Introduction}

Deep Learning (DL) based methods have achieved great success in a large number of applications such as image processing as they have been among the most powerful and accurate used techniques. For example, Deep Neural Networks (DNNs) can achieve high accuracy and performance on visual recognition or complex regression algorithms. However, in Neural Networks a large number of multiply-accumulate operations operations (MACs) and memory accesses are needed for the model's inference which is very energy-costly and time consuming \cite{DAN2021100520}. This computational complexity  combined with the inherent error resilience of DL led to a remarkable research in approximate DNN accelerators~\cite{axibm}.

The aim of approximate computing is to achieve significant savings in computational resources and memory. It involves reducing the model parameters or activations to a lower numerical precision using fixed-point arithmetic instead of the standard 32-bit floating point. Previous work has shown that an optimum bitwidth for the model's MAC operations can introduce negligible error \cite{yolofpga} while many accelerators use integer quantization \cite{googletpu} (i.e. INT8). Then, on the hardware side, performing approximate MAC operations (which mainly use integer arithmetic as well) can lead to high performance, power, and energy gains~\cite{axxdnnsurvey}.
For example, 35\% to 81\% energy savings are reported for less than 1\% accuracy loss~\cite{axxdnnsurvey}.
The diverse and vast space of approximate arithmetic units (e.g.,~\cite{evoapprox,alwann}) and their non trivial impact on the DNN accuracy, exacerbates the design complexity and, thus, the need for an approximate emulation framework becomes apparent.
Moreover, as DNNs become deeper, they become more sensitive to approximation~\cite{weightoriented} and hence, approximation-aware DNN fine-tuning is required to recover the error introduced by naive approximation (i.e. by just replacing exact multipliers with approximate ones) and achieve high inference accuracy~\cite{axxdnnsurvey}.
Popular DNN frameworks do not support approximate arithmetic because only libraries of accurate mathematical functions are inherently supported, thus emulation becomes extremely slow.

In this paper, we present AdaPT\footnote{AdaPT is available on the github repo: github.com/dimdano/adapt}, a fast emulation of approximate DNN accelerators in PyTorch that utilizes AVX intrinsics and multithreading. AdaPT aims to simplify and accelerate the process of simulating AI models for approximate hardware. It acts as a seamless PyTorch plugin that can be enabled for the majority of AI models such as Convolutional neural networks (CNNs), Variational autoencoders (VAEs), Long short-term memories (LSTM) and Generative adversarial networks (GANs). This is a novel emulation platform, first time enabling support for PyTorch whose ecosystem has been known for taking over AI researchers~\cite{stateai}. AdaPT can also perform approximate inference on any size of bitwidth representation supporting mixed precision as well. For the model quantization state-of-the-art techniques are used while approximate re-training or fine-tuning is also supported for further accuracy improvement.

\section{Related Work}

Popular frameworks for DNNs such as Caffe or TensorFlow have been extensively investigated by the community for approximate CNN simulation \cite{tfapprox, alwann} on image recognition. However, in the last years PyTorch has become the standard for DNN training or inference and to the best of our knowledge no previous work has shown support on this framework for approximate DNN emulation. Also, most of the previous works focus on 8-bit quantized CNNs only for image recognition simulation while AdaPT can support any bitwidth (e.g., 4bit, 8bit, 12bit, etc.) for many kinds of DNNs and applications such as CNNs for image recognition, LSTMs for text classification or VAEs and GANs for image reconstruction including support for approximation-aware re-training. 

Many pre-RTL simulation frameworks have been developed for approximate DNNs such as AxDNN \cite{10.1145/3287624.3287627} and TypeCNN \cite{8714855}. TypeCNN is evaluated on 2 Neural Networks (NNs) based on the Lenet architecture on a custom C++ framework while no CPU optimizations were employed. AxDNN combines precision-scaling and pruning methods with simulation of approximate hardware and it states $~20\times$ simulation speed up from the default RTL simulation for power analysis only. Other frameworks such Ristretto \cite{Gysel2016RistrettoHA} can evaluate various bitwidth representations but do not support approximate arithmetic. Additionally, ALWANN \cite{alwann} and TFApprox \cite{tfapprox} implemented different variations of ResNets using approximate units with 8-bit weights. ALWANN reported a very high simulation time ($\sim 1$ hour for ResNet50). Last, TFApprox similarly with ProxSiM \cite{proxsim} emulated the evaluation on a GPU achieving low inference time on Tensorflow framework but only limited to 8-bit inference. ProxSiM also had re-training capabilities for 8-bit multipliers but didn't show results on popular DNNs.

\section{Approximate computing in AdaPT}

We designed AdaPT framework for fast cross-layer DNN approximation emulation packaged as a PyTorch plugin which can be enabled by the user (or disabled if the PyTorch default flow is needed).
A plethora of layers and model architectures is seamlessly supported.
We support two major techniques for accuracy improvement; post training quantization using state-of-the-art calibration and approximate-aware retraining.
Also, the user can arbitrarily choose an approximate compute unit (ACU) to import in AdaPT as a black box or use the default accurate flow.
In addition, AdaPT supports mixed precision and mixed approximation (i.e., different ACU) in between layers.
Though, more fine-grained approximation (e.g., per filter) is not yet supported.
Last, the approximate DNN emulation is accelerated using OpenMP threads and vectorization with Intel AVX2 intrinsics which are advanced vector extensions.



\subsection{Optimized quantization}

In order to simulate approximate compute units we need first to perform an efficient quantization scheme that eliminates the error impact as much as possible. Previous work focused on 8-bit quantization \cite{tfapprox, proxsim} but in AdaPT we implement a generic bitwidth quantizer based on Nvidia's TensorRT toolkit \cite{toolkit} that can support lower or higher precision as well. For example, this can important for simulating higher precision ACUs for a variety of DNNs which do not have much error resilience, such as compact CNNs ~\cite{quant1, quant2, axxdnnsurvey}. We used this method as it is open-source and state-of-the-art, and proposing a new quantization method is out of the scope of the paper. Our quantizer maps a real number to an integer and can be applied on both weights and activations which are usually part of a Convolution or Linear layer. The mapping between real and quantized values must be affine, meaning they must follow the equation $real\_value = A \times quantized\_value + B$ where $A$ is the scale and $B$ is the zero point (often set to zero).

\subsubsection{Parameter calibration} 
 
In order to choose optimal quantization parameters for the scale values we used the \textit{calibrator} class from TensorRT to collect data statistics. We implemented the histogram calibrator for a 99.9\% percentile in our quantization modules as we saw it performed the best overall but other methods can be transparently used such as MSE (Mean Squared Error) or entropy.
So, instead of simply finding the max absolute number of our values, our calibrator learns offline $calib\_max$, the absolute maximum input value representable in the quantized space for a 99.9\% percentile.
Last, weight ranges are per channel while activation ranges are per tensor as previous work has also shown it performs well \cite{quant1}. Then, by only processing one batch of images our learnable $calib\_max$ can be configured optimally for most cases of DNNs (i.e. \textapprox 0.1\% error for most 8-bit CNNs).

\begin{figure}[t!]
\centering
  \includegraphics[width=\linewidth]{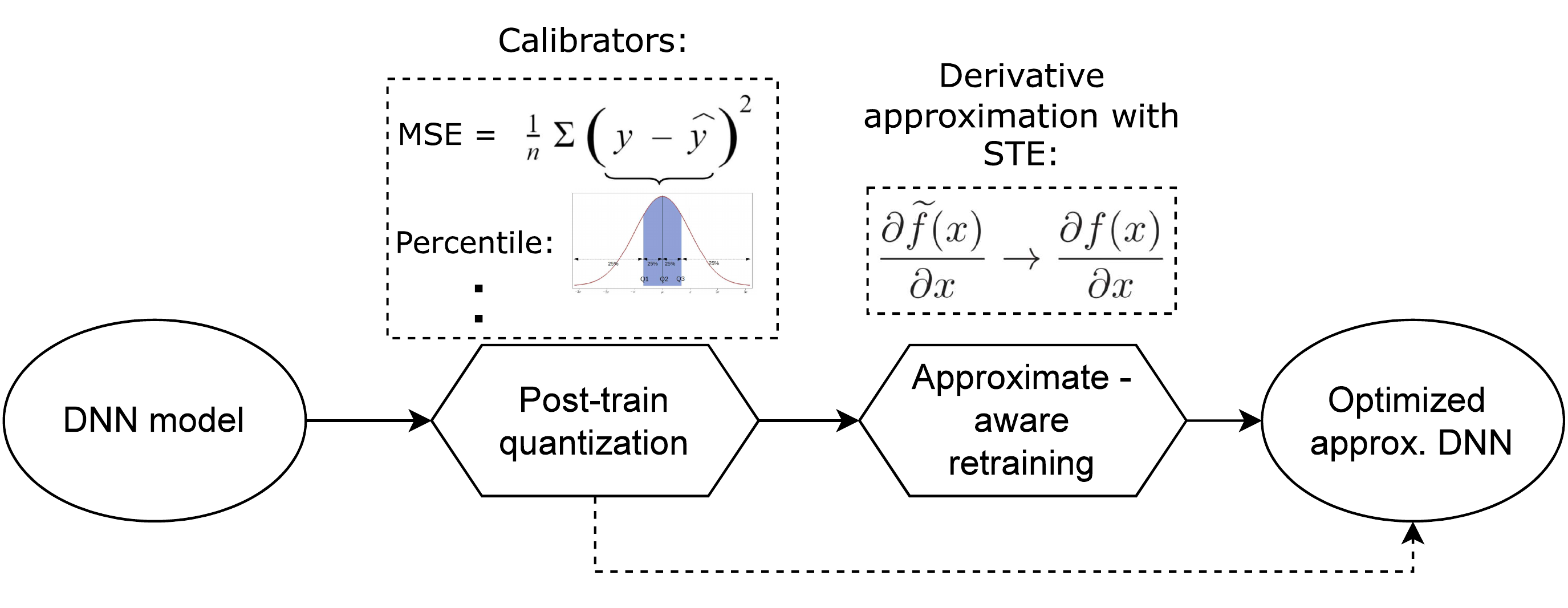}
  \caption{Quantization \& approximation-aware retraining flow before inference}
  \label{fig:train}
\end{figure}

\subsubsection{Retraining} 

Optionally, after post-training quantization we can perform Quantization Aware Training (QAT) by continuing training the calibrated model based on Straight Through Estimator (STE) derivative approximation. The process is illustrated in Figure \ref{fig:train}. AdaPT mitigates the effects of approximation during training by placing \textit{fake} quantization modules which work with quantized floating point values to simulate the rounding effects brought by true integer quantization, thus computing effectively the layer gradients. Last, during QAT the model performs propagation through our ACUs usually for 10\% of the default training schedule making the retraining \textit{approximation-aware}.

\subsection{Approximate units}

In AdaPT, the user can select in the definition of each DNN layer whether it is accurate or approximate.
Any arbitrary ACU can be specified for each layer (or all layers) as long as the multiplier’s output is deterministic. This section presents the most common layers that we re-designed and adapted for approximation.

\subsubsection{Convolution Layer}
  Usually for 2D convolution cases which often apply on CNNs we have an input of tensor X and shape $(N, C_{in}, H_{in}, W_{in})$, where N is the batch size, C is the number of channels, H is the height and W is the width, and an output of a tensor Y of shape $(N, C_{out}, H_{out}, W_{out})$. We expanded the filters into a 2-D matrix and the input matrix into another so that multiplying these 2 matrices would compute the same dot product as the original 2D convolution. The aim of this transformation is to allow a more efficient implementation for acceleration of AdaPT's emulation by simply computing a matrix multiplication. Our layer supports all kinds of input dimensions, kernel sizes, padding, striding and groups which enabled us to simulate many kinds of DNNs.
 
\subsubsection{Separable convolution} For separable convolution the main idea here is to transform it into a two-step calculation, depthwise and pointwise 2D convolution as in the equation:

 \begin{equation}\label{eq:quant}
\begin{aligned}
 y &= Conv2D(C_{in}, C_{in}, H_{in}, W_{in}, groups = C_{in}) (x) 
 \\
  o&ut = Conv2D(C_{in}, C_{out}, H'_{in}, W'_{in}, groups = 1) (y)
\end{aligned}
\end{equation}

The first equation comprises the depthwise convolution which is equivalent to a Conv2D with groups equal to  $C_{in}$ channels. Next the output is fed to the pointwise convolution which is same as Conv2D with $1\times1$ kernel size.

\subsubsection{Linear Layer}
Linear Layer is often found in MLPs, in the last layers of DNNs or even in GANs and VAEs. Similarly with 2D convolution, the PyTorch equivalent layer is a matrix multiplication of $y=xA^T+b$. The input matrix is multiplied with the weight matrix plus an optional bias vector. 

\subsubsection{RNN Layer}

Recurrent Neural Network (RNN) layers are typically used for temporal related tasks such as text classification or speech recognition. We implemented the feedback loop in the recurrent layer so that it maintains information in `memory' over time following an approach that is mathematically equivalent with the vanilla Pytorch RNN layer. It also utilizes our custom Linear layer thus making it \textit{approximation compatible} as well. Similarly, for Long short-term memory (LSTM) and Gated recurrent unit (GRU) layers we included the so called ``memory cell'' that can maintain information in memory for long periods of time. 

\subsection{Framework operation}

\begin{figure}[t!]
\centering
  \includegraphics[width=\linewidth]{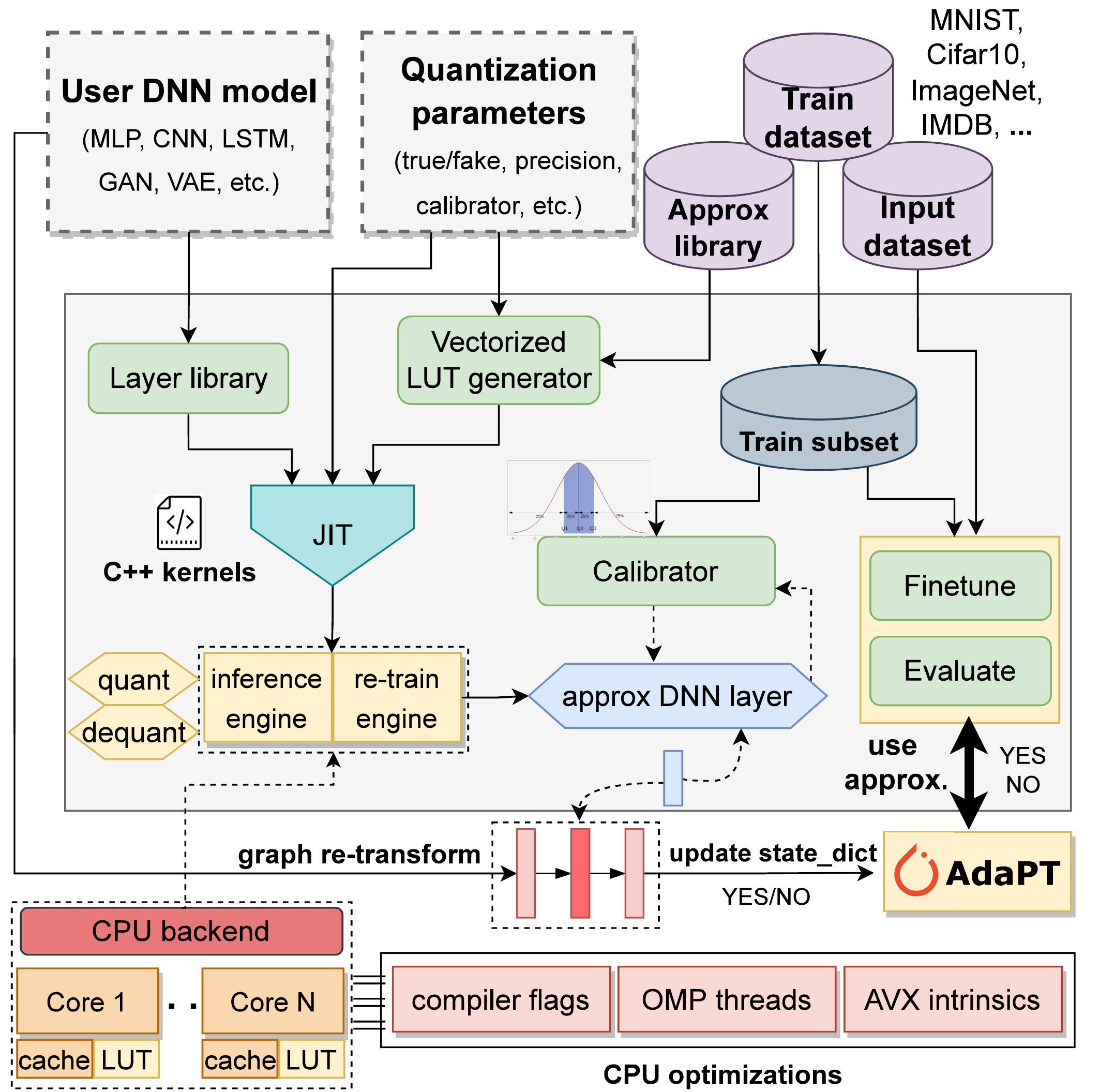}
  \caption{AdaPT framework operation}
  \label{fig:system}
\end{figure}

AdaPT framework operation is shown in Figure \ref{fig:system}. First, the user sets the desired DNN model with the quantization parameters needed such as precision used, calibrator, etc. Then, they define the approximate module to use from the library along with the dataset of the DNN models. It's worth mentioning that for the train dataset only a representative subset is needed which can be around 10\% of the original training set for the purposes of calibration. Then, AdaPT finds the supported layers in the DNN and fetches from its layer library the appropriate layer class.
Next, for the approximate multiplier, the corresponding LUT is produced from AdaPT's Look-up Table (LUT) generator in a cache-line aligned C-array which enables CPU cores to fetch data from the same cache \textit{chunk}. Also, a tool similar to our previous work~\cite{retsina} is used to translate a hardware description to a C function.
In case of large bitwidth where LUTs can increase substantially AdaPT can always substitute the LUT-based multiplication with functional-based multiplication (in which the approximate multiplier is alternatively described in C-code). This approach can alleviate the problem of memory in large LUTs (\textgreater 15bit) but can introduce overhead in the DNN execution time. It's worth mentioning that both approaches provide a 1-1 representation of the ACU at high-level thus the results would be the same in the quantization or re-training.
Generally, AdaPT tries to populate the CPU cores cache with the LUTs as much as possible in order to minimize cache misses. Last, just in time (JIT) compilation loads the layer extension on the fly using Ninja build system which builds the sources. The produced inference and retrain engines are linked with the final approximate DNN layers which will substitute for the corresponding vanilla PyTorch layers using a graph re-transform tool. The tool analyses the layers and recursively changes the PyTorch layers with the approximate equivalents. Finally, user can optionally fine-tune the model using the provided train subset in order to achieve even higher accuracy or just proceed with the approximate evaluation.

\section{Accelerated emulation in AdaPT}

The most computationally intensive part of the implemented layers is based on matrix multiplications. For example, the original 2D convolution is transformed to matrix multiplication. 
To this end, in order to support approximate units we had to implement the inner multiplications between each input and filter values using a LUT. Thus, we would compute any approximate unit without the need to implement its corresponding function directly (except for the cases of large LUT sizes as mentioned earlier). Then, the table look-ups are parallelized in a hybrid model of parallel programming using OpenMP threads and CPU vectorization.



\subsection{Thread Parallelism}
 In AdaPT we created an efficient batched Conv2D implementation which scales with input size and does not suffer from memory errors using the thread parallelism of OpenMP. It enabled us to perform loop-based parallelism shared between batches and achieve almost linear scaling when input data size grows. The main goal of the aforementioned approach is to unify the usage of gradual parallelism through a common interface and ease the application development.
 
\begin{figure}[t!]
\centering
  \includegraphics[scale=0.5]{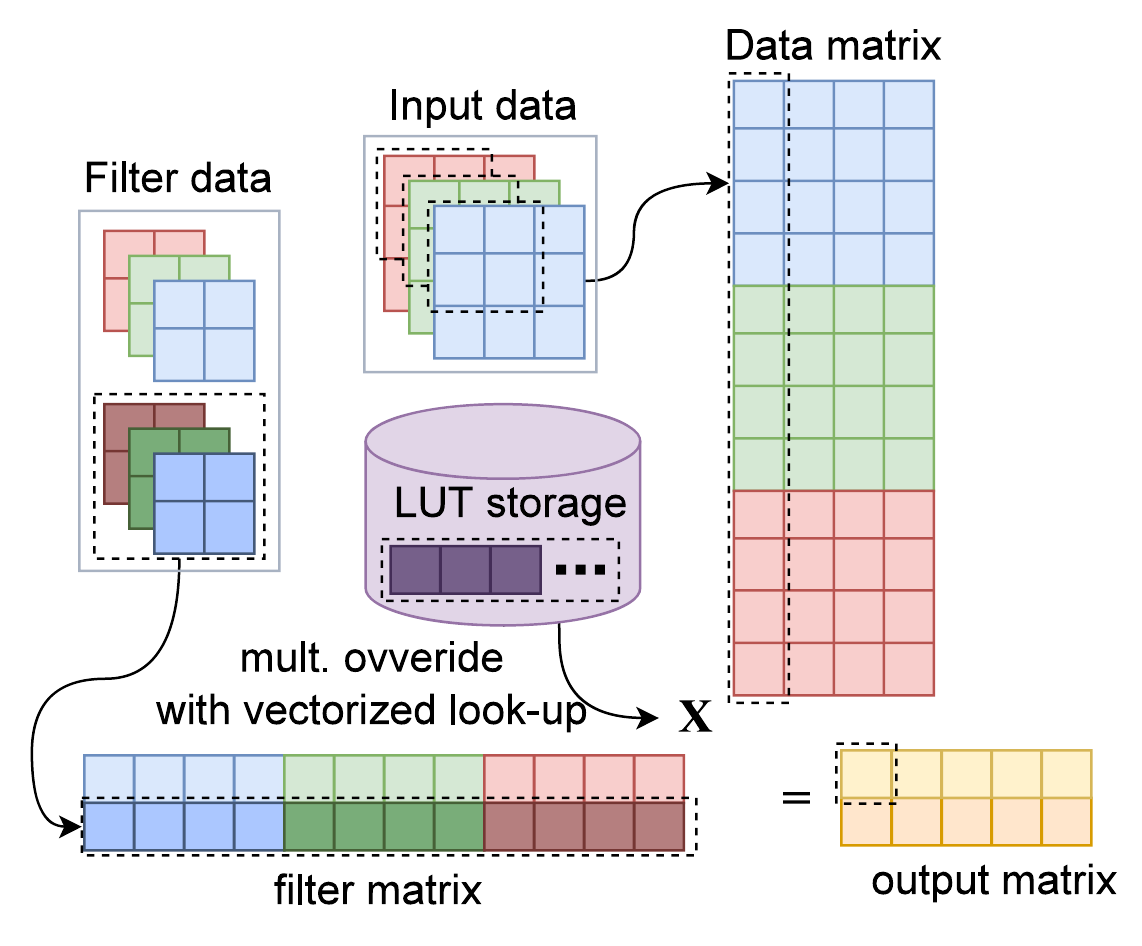}
  \caption{2D convolution to matrix multiplication with vectorized LUT override}
  \label{fig:gemm}
\end{figure}

\subsection{Vector Parallelism}

The second level of parallelism is introduced on each thread's execution which relates to the implementation of parallel table lookup using Single Instruction Multiple Data (SIMD) instructions. The aim is to accelerate the \textit{gather} operation of the look-up table data which is the process of the data taken from disjoint locations in memory and stored in continuous memory. Towards allowing efficient loads from memory, all values of the SIMD need to be in contiguous memory which is the case for our AdaPT tensors. The indices which are the activations combined with the weights are packed into vector registers. The AVX2 instruction set can be then utilized to implement the gather instruction and vectorize the task by gathering the memory locations from the LUT into the destination vector register. We chose to use AVX2 intrinsics due to their broad support in Intel CPUs shipped in 2013 and later and their support in AMD CPUs as well. In Figure \ref{fig:gemm} we illustrate the process of using vectorized loads in a 2D convolution scenario.



\section{Evaluation and Results}

In this section we present the evaluation of AdaPT regarding several DNN networks with their respective quantization calibration, approximation-aware training and simulation time. 
The experiments were conducted on Intel Xeon Gold 6138 CPU at 2.00GHz and 64GB RAM.
We evaluated different scenarios of applications such as image classification with CNNs on Cifar10 and ImageNet datasets, text classification with LSTM on IMDB dataset, and last image reconstruction with GAN and VAE on Fashion MNIST and MNIST dataset respectively.
Below, in Table \ref{tab:models}, we summarize the specifications for each model used in the experiments regarding the corresponding type, the dataset used for evaluation/retraining, number of model parameters (Params) and operations (OPs).

\begin{table}[t!]
\centering
\footnotesize
  \caption{Specifications for each DNN used in our experiments}
  \label{tab:models}
  \begin{tabular}{l|cccc}
    \toprule
    DNN & Type & Dataset & Params  & OPs  \\
    \midrule
    ResNet50 & CNN & CIFAR10 & 23.52M  & 0.33G \\
    DenseNet121 & CNN & CIFAR10 & 6.96M  & 0.23G \\
    VGG19 & CNN & CIFAR10 & 38.86M  & 0.42G \\
    Fashion-GAN & GAN & Fashion MNIST & 0.28M  & 0.29M \\
    VAE-MNIST & VAE & MNIST & 0.65M & 0.66M \\
    LSTM-IMDB & LSTM & IMDB & 0.58M  & 0.55G \\
    Inceptionv3 & CNN & ImageNet & 27.16M  & 2.85G \\
    SqueezeNet & CNN & ImageNet & 1.24M  & 0.36G \\
    ShuffleNet & CNN & ImageNet & 2.28M  & 0.15G \\

  \bottomrule
\end{tabular}
\vspace{2ex}
\centering
\footnotesize
  \caption{Accuracy and retrain time evaluation on various DNNs}
  \label{tab:model_acc}
  \setlength{\tabcolsep}{2.2pt}
  \begin{threeparttable}
  \begin{tabular}{l|cccc|cc}
    \toprule
    \multicolumn{7}{l}{mul8s\_1L2H \ \ \  MAE: 0.081 \%, MRE: 4.41 \%, power: 0.301mW\tnote{1}} \\
    \midrule
    DNN & FP32 & 8bit & 8bit calib. & 8bit approx. & retrain\tnote{3} & time \\
    \midrule
   ResNet50 & 93.65\% & 93.55\% & 93.59\%& 82.69 \% & 93.44\% & 763s\\
    VGG19 & 93.95\% & 93.80\% & 93.82\%& 90.7\% & 93.56\% & 318s\\
    VAE-MNIST & 99.99\% & 99.95\% & 99.96\%& 93.12\%& 99.88\%& 9.28s\\
    LSTM-IMDB & 83.10\% & 82.90\% & 82.95\%& 79.9\% & 82.63\% & 710s\\
    SqueezeNet & 80.6\% & 79.01\% & 80.16\%& 62.01\% & 76.21\% & 620s\\
    \midrule
    \midrule
    \multicolumn{6}{l}{mul12s\_2KM \ \ \  MAE: 1.2e-6 \%, MRE: 4.7e-4 \%, power: 1.205mW\tnote{2}} \\
    \midrule
    DNN & FP32 & 12bit & 12bit calib. & 12bit approx. & retrain\tnote{3} & time \\
    \midrule
   ResNet50 & 93.65\%  & 93.60\% &93.61\%&  93.52\%& 90.54\% & 798s\\
    VGG19 & 93.95\%  & 93.80\% & 93.81\%& 93.81\% & 93.71\% & 359s\\
    VAE-MNIST & 99.99\% & 99.98\% & 99.98\%& 99.98\%& 99.99\%& 10.11s\\
    LSTM-IMDB & 83.10\% & 82.94\% & 82.96\%& 82.96\% & 83.12\% & 1040s\\
    SqueezeNet & 80.6\%  & 80.11\% & 80.3\%& 80.35\% & 80.50\% & 623s\\
    
  \bottomrule
\end{tabular}
  \begin{tablenotes}\footnotesize
\item[] $^1$power of 8bit exact: 0.425mW. $^2$power of 12bit exact: 1.210mW.\\
$^3$ approx. multiplier \& approximation-aware retrain.
\end{tablenotes}
\end{threeparttable}
\end{table}



\subsection{Quantization and retraining}

Five metrics are evaluated for each model: the accuracy in the default FP32 models, quantized models (with and without calibration), approximate models, and the models taken after approximate-aware re-training. Towards post-quantization calibration we used only two batches of images which we set as 128 in order to collect the histograms (based on a 99.9\% percentile method). We retrained the DNNs through Stochastic Gradient Descent (SGD) with a learning rate of 1e-4 and a batch size of 128. For the retrain subset we used 10\% of the corresponding training datasets which was adequate to fine-tune our models. For demonstration of the retraining, we used five representative DNNs for image recognition (ResNet50, VGG19, SqueezeNet), text classification (LSTM-IMDB) and image reconstruction (VAE-MNIST) tasks. The aforementioned DNNs were tested with two approximate multipliers (but any can be used since they are all implemented as LUTs) with distinct Mean Relative Error (MRE) and Mean Absolute Error (MAE) values taken from EvoApprox library \cite{evoapprox}. One with 8-bit precision and low power consumption but higher MRE and one with 12-bit precision with lower MRE but higher power consumption. Last, top-1 accuracy metric was used in general except for ImageNet models which used top-5. 

The aforementioned results, observed in Table \ref{tab:model_acc}, show that post-training quantization attains low accuracy error from the original FP32 models ($\sim 0.1\%$) due to calibration. Generally, calibration is important for moden neural networks especially larger ones.
More information regarding the impact of calibration can be found in~\cite{calibration}.
With our approximate-aware retraining we can \textit{adapt} DNNs to the custom approximate backward engine in order to deliver higher accuracy to the approximated DNN. According to previous work \cite{one-epoch}, we also retrained most of the models for a single epoch (i.e. SqueezeNet was trained for 14 as typically Imagenet models require more epochs) achieving significant performance. Also, the effectiveness of AdaPT's retrain engine can be evident from the substantial error recovery of approximation, especially in the 8bit ACU ($\sim 7.5\%$ increase on average). The error can be reduced but very slightly by more fine-tuning with learning rate annealing.

\begin{table}[t]
  \caption{Inference emulation time for different DNNs}
  \label{tab:results}
  \footnotesize
  \setlength{\tabcolsep}{2pt}
  \begin{tabular}{l|lll|lc}
    \toprule
    DNN & Native CPU & \makecell{ Baseline \\ Approx. }& \makecell{ AdaPT \\ (w/ func.)} & \makecell{ AdaPT \\ (w/ LUT)} &  \makecell{ AdaPT \\ vs Baseline} \\
    \midrule
    ResNet50  & 0.5 min & 76.5 min & 104 min & 1.7 min  & 45x \\
    DenseNet121  & 0.48 min & 53.2 min & 72 min & 1.6 min  & 33.2x  \\
    VGG19 & 0.2 min & 91.7 min & 125 min & 1.7 min & 53.9x \\
    Fashion-GAN  & 0.003 min & 0.02 min & 1.1 min & 0.012 min  & 1.7x\\
    VAE-MNIST  & 0.015 min & 0.1 min & 1.2 min & 0.02 min  & 5x\\
    LSTM-IMDB & 1.36 min & 48.5 min & 449 min & 7.6 min  & 6.4x\\
    Inceptionv3  & 22.1 min & 2909 min & 4560 min & 83 min &  35.1x \\
    SqueezeNet  & 11.6 min & 443 min & 576 min & 20.6 min &  21.5x \\ 
    ShuffleNet  & 11.4 min & 163 min & 251 min & 22.4 min &  7.3x \\
    \bottomrule
  \end{tabular}
\vspace{3ex}
    \centering
  \caption{Qualitative comparison with state-of-the-art}
  \label{tab:tool_support}
  \footnotesize
  \setlength{\tabcolsep}{3pt}
  \renewcommand{\arraystretch}{1.1}
  \begin{threeparttable}
  \begin{tabular}{l|c|cccc}
    \toprule
    
 Tool Support & AdaPT & \cite{tfapprox} & \cite{proxsim} & \cite{alwann} &  \cite{8714855}
 \\
    \midrule
     Framework & PyTorch & TF\tnote{1} & TF & TF & C++ \\
     Backend & CPU & GPU & GPU & CPU & CPU\\
     Varying DNN types\tnote{2} & \cmark & \xmark & \xmark & \xmark & \xmark\\
     Arbitrary ACU & \cmark & \xmark & \xmark & \xmark & \cmark\\
     Quantization calibration & \cmark & \xmark & \xmark & \cmark & \xmark \\
     Approximate-aware retraining & \cmark & \xmark & \cmark & \cmark & \cmark\\ \hline

  \end{tabular}
  \begin{tablenotes}\footnotesize
\item[] $^1$TF: Tensorflow. $^2$For example: CNN, LSTM, GAN, etc.
\end{tablenotes}
\end{threeparttable}
\vspace{-2ex}
\end{table}

\subsection{Inference emulation}

In this section we summarize the emulation time for each approximate DNN on Table \ref{tab:results}. Additionally, performing quantization and dequantization in each layer introduces some overhead around 10\% for the optimized approximate solution.
The inference comparison is done on 8-bit in accordance with the related work for unbiased comparison (the approximate module can be arbitrary since all are implemented as LUTs).
However, the inference time is lower when using small LUTs due to better cache usage (also we observed a $\sim 2.1\times$ increase in time on average when expanding the LUT bitwidth by two). Last, we compare AdaPT with PyTorch \textit{native} FP32 optimized implementation, the \textit{baseline} unoptimized approximate simulation which uses LUTs but omits our optimizations and with the functional C-implementation of the ACU (\texttt{mul8s\_1L2H}). 

In Table \ref{tab:results} we can observe the computation time for AdaPT was greatly reduced compared with the baseline approach. When compared with the state-of-the-art, it is significantly lower than ALWANN \cite{alwann} which runs also on a Xeon CPU (1.7min vs 54.5min on ResNet50).
TypeCNN \cite{8714855} which runs on CPU does not report inference results while it runs on a custom C++ framework. Next, when compared with ProxSim~\cite{proxsim} the execution time is very similar (20.6min vs 17.5min on SqueezeNet) despite running on a GPU. TFApprox \cite{tfapprox} runs faster on a GPU with ResNet50 (1.7min vs 0.26min) but the authors only examine/support ResNets on 8-bit inference for image recognition. The diversity of AdaPT features such as supporting a variety of model architectures, application domains, approximations or approximation-aware retraining compose a very robust framework. A comparison of AdaPT's functionalities with state-of-the-art is shown in Table \ref{tab:tool_support}. 


\section{Conclusion and Perspective}

In this paper we presented AdaPT, an end-to-end framework for fast cross-layer evaluation and re-training of approximate DNNs based on the popular PyTorch library. AdaPT simplifies and accelerates the process of DNN simulation using multi-threading and vectorization while at the same time it can support a wide range of DNN topologies for various deep learning tasks such as image recognition, text classification and image reconstruction. Through diverse experiments, we demonstrated the \textit{adaptivity} of our framework with various DNNs, ACUs, and application domains and paved the way to new approximate DNN accelerators first time for PyTorch. 

\bibliographystyle{IEEEtran}
\bibliography{references}

\end{document}